%% file: paper.tex
\documentclass{article}

\usepackage{microtype}
\usepackage{graphicx}
\usepackage{subfigure}
\usepackage{booktabs} 
\usepackage{xcolor} 
\usepackage{listings}
\usepackage[toc,page]{appendix} 
\usepackage{kotex}

\usepackage{enumitem}
\setlist{nolistsep,leftmargin=*}

\usepackage{hyperref}

\usepackage[accepted]{sysml2019}

\sysmltitlerunning{CHOPT : Automated Hyperparameter Optimization Framework for Cloud-Based Machine Learning Platforms}

\begin{document}

\twocolumn[
\sysmltitle{CHOPT : Automated Hyperparameter Optimization Framework for Cloud-Based Machine Learning Platforms}



\sysmlsetsymbol{equal}{*}

\begin{sysmlauthorlist}
\sysmlauthor{Jinwoong Kim}{equal,na}
\sysmlauthor{Minkyu Kim}{equal,na}
\sysmlauthor{Heungseok Park}{na}
\sysmlauthor{Ernar Kusdavletov}{uni}
\sysmlauthor{Dongjun Lee}{na}
\sysmlauthor{Adrian Kim}{na}
\sysmlauthor{Ji-Hoon Kim}{na}
\sysmlauthor{Jung-Woo Ha}{na}
\sysmlauthor{Nako Sung}{na}
\end{sysmlauthorlist}

\sysmlaffiliation{uni}{Department of Computer Science, Ulsan National Institute of Science and Technology, Ulsan, Korea}
\sysmlaffiliation{na}{Clova AI Research, NAVER Corp., Seongnam, Korea}

\sysmlcorrespondingauthor{Jung-Woo Ha}{jungwoo.ha@navercorp.com}


\vskip 0.3in

\begin{abstract}

Many hyperparameter optimization (HyperOpt) methods assume restricted computing resources and mainly focus on enhancing performance. 
Here we propose a novel cloud-based HyperOpt (CHOPT) framework which can efficiently utilize shared computing resources while supporting various HyperOpt algorithms. 
We incorporate convenient web-based user interfaces, visualization, and analysis tools, enabling users to easily control optimization procedures and build up valuable insights with an iterative analysis procedure. 
Furthermore, our framework can be incorporated with any cloud platform, thus complementarily increasing the efficiency of conventional deep learning frameworks. 
We demonstrate applications of CHOPT with tasks such as image recognition and question-answering, showing that our framework can find hyperparameter configurations competitive with previous work. 
We also show CHOPT is capable of providing interesting observations through its analysing tools.
\end{abstract}
]



\printAffiliationsAndNotice{\sysmlEqualContribution} 

\input{1_introduction}
\input{2_related_work.tex}
\input{3_automl.tex}
\input{4_automlvis}
\input{5_usecase}
\input{6_evaluation}
\input{7_conclusion}

\bibliography{style/paper}
\bibliographystyle{sysml2019}

\end{document}

%% file: 1_introduction.tex
\section{introduction}~\label{sec:introduction}
Deep neural networks (DNNs) have become an essential method for solving difficult tasks in computer vision, signal processing, and natural language processing~\cite{he2016deep,StarGAN2018,han2017deep,van2016wavenet,seo2016bidirectional, vaswani2017attention}.
As the capabilities of deep learning have expanded with more modular architectures and advanced optimization methods, the number of hyperparameters has increased in general. This increase of hyperparameter sizes makes it more difficult for a researcher to optimize a model, wasting a lot of human resources and potentially leading unfair comparisons.
This reinforces the importance of efficient automated hyperparameter tuning methods and interfaces.

To address this problem, several hyperparameter optimization (HyperOpt) methods 
have been proposed~\cite{jaderberg2017population,falkner2018bohb,li2017hyperband}.
These methods have many advantages such as strong final performance, 
parallelism, early stopping which significantly improve performance in terms of computing 
resource efficiency and optimization time. 
However, for using the previous methods, users must implement it in their own system or code, and even if the source code is provided, considerable time and efforts are required for implementation.


Recently, there have been proposed HyperOpt frameworks to eliminate this integration cost ~\cite{liaw2018tune,tsirigotis2018orion,golovin2017google}. These frameworks host methods of optimizing hyperparameters and provides a visualization tool for analyzing the results of a trained model. However, the proposed frameworks do not consider computing resource management. Visualization tool plots only one type for whole parameter information but not supports interactive function to analyze properties and trends between user-specified hyperparameters.

In this work, we 
propose an efficient HyperOpt framework called Cloud-based Hyperparamter OPTimziation (CHOPT). 
CHOPT efficiently improves the resource usages in flexible computing environments based 
on \textit{Stop-and-Go} approach, with supporting state-of-the art hyperparameter optimization 
algorithms. Our CHOPT framework also provides web-based analytic visual tool which users can monitor the current tuning progress of models or compare conducted results with other ones.
Moreover, user can fine tune hyperparameters through visualization tool.

The main contributions of this work are as follows
\begin{itemize}
    \item We develop an automated HyperOpt framework that hosts hyperparamter optimization methods. We have shown that our framework can find competitive hyperparameters compared to results reported in previous papers.
    \item We propose \textit{Stop-and-Go} to address the early stopping problem and maximize utilization of shared-cluster resources, which we demonstrate in our evaluation.
    \item We introduce an implemented analytic visual tool and show a step by step use case of hyperparameter fine tuning with a real world example.
\end{itemize}


%% file: 2_related_work.tex
\section{related work}~\label{sec:related_work}




\subsection{Hyperparameter Optimization}

AutoML is a general concept which covers diverse techniques for automated model learning including automatic data preprocessing, architecture search, and model selection. 
HyperOpt is one of core components of AutoML because hyperparameter tuning has large influence on performance for
a fixed model structure even, in particular in neural network-based models. We mainly focuses on automated HyperOpt throughout this paper.


To this end, there have been proposed several hyperparameter optimization
algorithms ~\cite{jaderberg2017population,falkner2018bohb,li2017hyperband}.
Population-Based Training (PBT)~\cite{jaderberg2017population} is an
asynchronous optimization algorithm which effectively utilizes a fixed
computational budget to jointly optimize a population of models and their
hyperparameters to maximize performance. PBT discovers a schedule of
hyperparameter settings rather than following the generally sub-optimal
strategy of trying to find a single fixed set to use for the whole course of
training.
In practice, Hyperband~\cite{li2017hyperband} works very well and typically
outperforms random search and Bayesian optimization methods operating on the
full function evaluation budget quite easily for small to medium total budgets.
However, its convergence to the global optimum is limited by its reliance on
randomly-drawn configurations, and with large budgets its advantage over random
search typically diminishes.  BOHB~\cite{falkner2018bohb} is a combination of
Hyperband and Bayesian Optimization. Hyperband already satisfies most of  the
desiderata (in particular, strong anytime performance, scalability, robustness
and flexibility), and with BOHB it also satisfies the desideratum of strong
final performance.

Although these hyperparameter optimization methods outperform traditional
methods in terms of resource utilization and optimization time, it requires
high cost to apply on a new environment or code. 
Moreover, it is hard to know which solution performs better than others before
applying all methods to the model. For this reason, there have been some
efforts to solve these problems through systematic improvements.



\subsection{Hyperparameter Optimization Framework}

Tune~\cite{liaw2018tune} is a scalable hyperparameter optimization framework
for deep learning that provides many hyperparameter optimization methods.
However, Tune requires user code modification to access intermediate training
results while our framework does not require any user code modification. Also,
Tune provides visual tool via web interface, but user cannot interfact with it
for fine tune, analyzing, or changing hyperparameter sets.
Or{\'\i}on~\cite{tsirigotis2018orion} is another hyperparameter optimization
tool which mainly focuses on version control for experiments. It is designed to
work with any language or framework since it has a very simple configuration
interface, however, it does not really manage computing resources such as GPUs
to improve performance at optimization time and resource efficiency. Google
Vizier~\cite{golovin2017google} is a Google-interval service. Although it
provides many hyperparameter optimization, parallel execution, early stopping
and many other features, it is closed-source project so it is hard to apply
this system to user's code or system.

\subsection{NSML}
NSML~\cite{sung2017nsml} is a platform designed for helping machine learning
developers and researchers.  Although many machine learning libraries such as
TensorFlow~\cite{abadi2016tensorflow} and PyTorch~\cite{paszke2017automatic}
make many researcher's life easy by simplifying model implementation, they
still suffer from a non-trivial amount of manual tasks such as computing
resource allocation, training model progress monitoring, and comparison of
models with different hyperparameter settings.

In order to handles all these tasks so that the researchers can focus on
models, NSML includes automatic GPU allocation, learning status visualization,
and handling model parameter snapshots as well as comparison of performance
metrics between models via a leaderboard.

We chose NSML as a cloud machine learning platform for implementing and evaluating the proposed CHOPT due to the following properties: 1) since NSML has
well-designed interfaces including Command Line Interface (CLI) and web, so that
it is easy to assign the hyperparameters to model and get the result from the model.
2) NSML enables us to easily manage cluster computing resources and effectively monitor the progress of model training.

%% file: 3_automl.tex
\section{CHOPT: Cloud-based Hyperparameter OPTimizaion}~\label{sec:automl}

\subsection{Design goals and Requirements}
The main goals of designing our framework are as follows.

 
\textbf{Efficient resource management.}
We need our framework to manage shared resources efficiently so that it balances throughout CHOPT users and non-CHOPT users while maximizing resource utilization.

\textbf{Minimum configuration and code modification.}
By hosting hyperparameter optimization methods in our system, users should be able to easily tune their models without implementation overheads.
In addition, configurations should be simple yet flexible.

\textbf{Web-based analytic visual tool.}
As hyperparameter optimization requires many training sessions, it is essential to have powerful visualization tools to analyze from hundreds of training results.
 
\subsection{System Architecture}

\begin{figure}[H]
\centering
\includegraphics[width=0.45\textwidth]{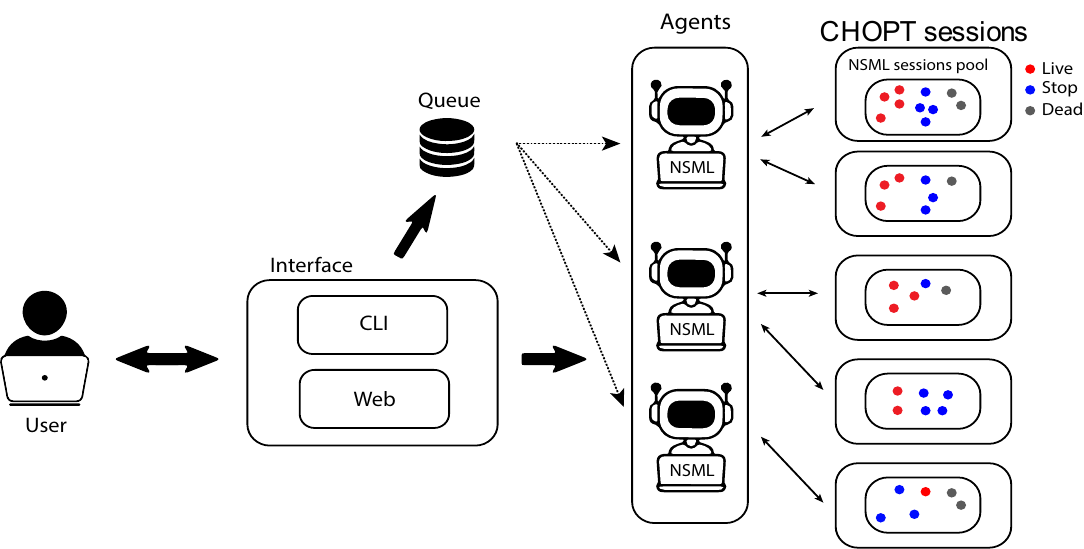}
\caption{System Architecture of CHOPT}
\label{fig:arch}
\end{figure}



In this section, we would like to introduce our hyperparameter optimization system step-by-step, where Figure \ref{fig:arch} is the architecture of the proposed CHOPT framework.

Users can run \textbf{CHOPT sessions} through both a command line interface (CLI) or Web interface.
CLI is light-weight, therefore suitable for simple tasks such as running and stopping sessions.
The web interface is relatively heavy but provides many features, making it more suitable for analyzing and monitoring training results.
To run a CHOPT session, the user needs to submit codes compatible with NSML and a configuration file containing details on the tuning method, hyperparameter sets, and more.
While holding the submitted codes, a CHOPT session is initilized with the configuration file and gets inserted in a \textbf{Queue}, which stores CHOPT sessions before running.
When a CHOPT session manager (\textbf{Agent}) is available, a CHOPT session obtained from the Queue is ran by the Agent.

\subsubsection{Agent}
An Agent is a module responsible for running CHOPT sessions as well as monitoring the progress of an automl session so that users can check and analyze corresponding NSML training sessions with given interfaces. Note that NSML session is a single training model.


To manage and maximize computing resource CHOPT sessions have three session pools, \textit{live pool, stop pool,} and \textit{dead pool}.
All running NSML sessions are in the live pool, which have resource limits set by the configuration file.
With NSML session in live pool, agent periodically compares the performance of NSML sessions and tuning NSML sessions according to the configuration file.
If NSML session is exited, it will be moved to stop pool or dead pool.
NSML session in stop pool can be restarted while NSML session in dead pool is completely removed from our system since automl systems commonly creates models a lot and it often takes up too much system storage space. 
Thus, user can specify the \textit{stop ratio}, how much of exited NSML session goes to stop pool and then NSML session in stop pool can be resumed when system becomes under-utilized or recovered by user. 

\subsubsection{Master Agent}
Master agent assigns a session to one of agents according to the configuration file.
Master agent is elected from one of agents like zookeeper's leader 
election~\cite{hunt2010zookeeper}. If master agent falls, any agent 
can be the next master agent. Whenever user submits CHOPT session into the queue,
One of the most important roles of the master agent is shifting available computing
resources such as GPUs between CHOPT users and Non-CHOPT users. Otherwise,
users will face serious computing resource imbalances. We describe the details of this in
the next Section~\ref{sec:resource_management}.

\subsection{Stop-and-Go}~\label{sec:resource_management}
Automated optimization methods and frameworks often take a lot of computing resources because many models are trained in parallel.
If not carefully managed, this leads to crucial load imbalance across users in shared cluster environments.
To avoid this problem, our framework controls available computing resources such as GPUs among
CHOPT users and non-CHOPT users with fairness while maximizing the utilization
of computing resources.
Here, we introduce a new feature called \textit{Stop-and-Go}.
The key idea of this feature is that the master agent controls available resources among NSML and CHOPT sessions according to the cluster's condition.
By doing so, we can maximize the utilization of shared cluster environment as well as mitigate 
the problem of early stopping in CHOPT environments.

\subsubsection{Efficient Resource Management}
Whenever a resource cluster is under-utilized, the master agent assigns more resources (GPUs) to CHOPT sessions so that they can quickly finish hyperparameter optimization.
On the other hand, if cluster is over-utilized, the master agent takes GPUs from CHOPT sessions so that other non-CHOPT users can train their models on the shared-cluster environment.
This feature enables to maximimum utilization of computing resources and efficiently train models.

\subsubsection{Resolving Early Stopping Problem}
Many hyperparameter optimization methods and frameworks have a feature called,
\textit{Early Stopping}~\cite{liaw2018tune,golovin2017google,jaderberg2017population,li2017hyperband}
which terminates unpromising model training earlier rather than keeps them
training. This feature can significantly save computing resources and also
enables researchers to train more models during the same time.

\begin{figure}[h]
\centering
\includegraphics[width=0.35\textwidth]{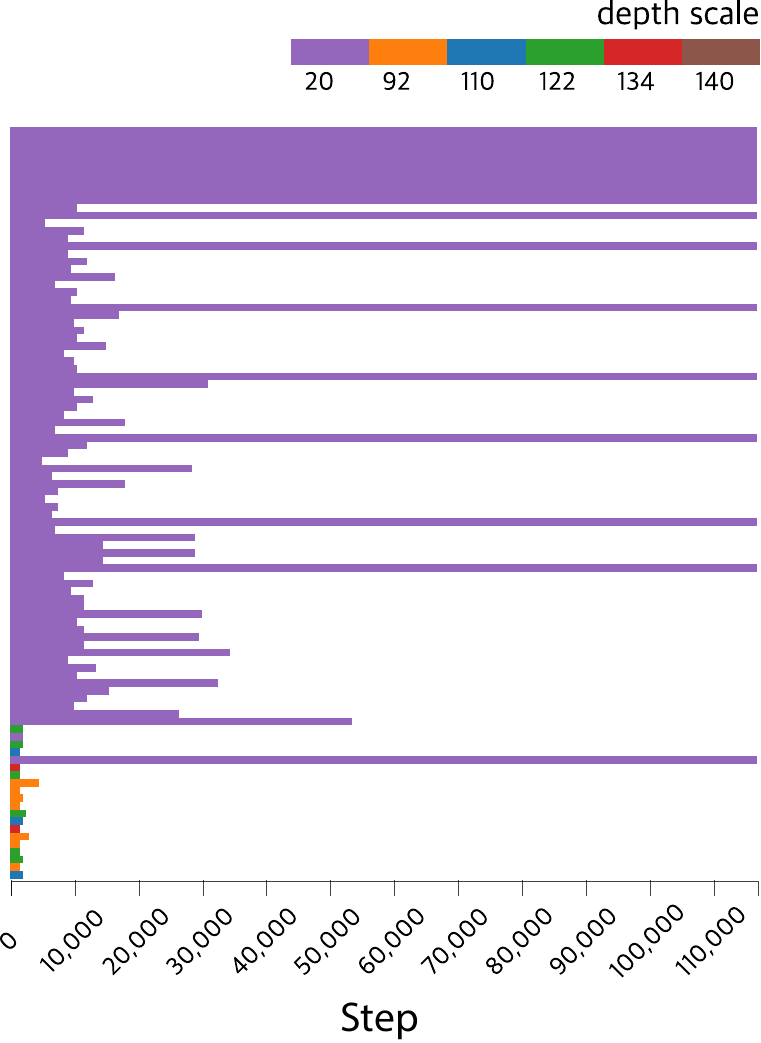} 
\caption{Hyperparameter Optimization with Early Stopping. Colors indicate the depth of the model as a hyperparameter to showcase the negative effects of early stopping.}
\label{fig:early_stopping_barPlot}
\end{figure}

However, early stopping can be harmful since models with high performance are not guaranteed to train quickly.
Figure~\ref{fig:early_stopping_barPlot} presents a history of model search by our framework with step size of 7 epochs.
Here, we refer step size as the checking interval for early stopping.
After a few steps, CHOPT with early stopping only gets to search a space with shallow depth because models with shallow depth perform better in the early stage of training than others.
This may cause a premature convergence of CHOPT, ending up with poor models.
More importantly, it is difficult to avoid because setting a small number of epochs as a step size would be the first option to choose as a beginner of CHOPT.

The proposed Stop-and-Go method can help users avoid the pitfall.
In our framework, when the Master agent takes GPUs from a CHOPT session, the session randomly splits running NSML sessions into the stop pool and dead pool.
If the CHOPT session gets allocated with more GPUs, it attempts to resume NSML sessions from the stop pool instead of creating new sessions, which is only done when the stop pool is empty.
By reviving early stopped sessions which have potential, we can avoid the problem of early
stopping.

\subsection{Configuration}~\label{sec:configuration}

\begin{table*}[ht]
\begin{lstlisting}[caption=Example of configuration]
config = {
    'h_params': {
        'lr': {'parameters': [0.01, 0.09], 'distribution': 'log\_uniform', 
               'type': 'float', 'p_range': [0.001, 0.1]},
        'depth': {'parameters': [5, 10], 'distribution': 'uniform', 'type': 'int', 
                  'p_range': [5, 10]},
        'activation': {'parameters': ['relu', 'sigmoid'], 'distribution': 'categorical', 
                       'type': 'str', 'p_range': []}
    },
    'h_params_conditions': [],
    'h_params_conjunctions': [],
    'measure': 'test/accuracy',
    'order': 'descending',
    'step': 5,
    'population': 5,
    'tune': {'pbt': {'exploit': 'truncation', 'explore': 'perturb'}},
    'termination': {'max_session_number': 50}
}
\end{lstlisting}
\label{code:config}
\end{table*}

Many hyperparameter optimization frameworks require users to modify their code.
However, it takes time to understand and perform implementations for these modifications.
In contrast with other frameworks, CHOPT does not require any code modification while require minimum configuration to tune their models on our framework. 
In this section, we introduce the typical form of our framework's configuration file, which uses a dictionary based structure.
We provide many examples of configuration files in our system for user convenience. 

\subsubsection{Hyperparmeter Space Definition}
Defining hyperparameter space is one of the most important steps in automated machine learning.
Hyperparameter space should be large enough to cover many points, and it should be small enough to search points efficiently at the same time.
We chose to use a dictionary format from python for hyperparameters because python is one of the most popular programming language among ML researchers.
Code in Listing 1 shows an example of an interface for hyperparameter space which has four components; parameters, distribution, type, and range.
The framework supports various distributions other than uniform such as Gaussian distributions.
With the simple python-dictionary-style configuration, users can easily add or remove any hyperparameters, and also compare two CHOPT sessions with different hyperparameter configurations via web interface.
As deep learning models get more complex to get better performance, in many cases they require hierarchical hyperparameters.
For instance, if a model uses stochastic gradient descent~\cite{ruder2016overview} as an optimizer, training would require momentum and learning rate as hyperparameters.
To fulfill these requirements, CHOPT also supports hierarchical hyperparameter space.

\subsubsection{CHOPT Session Configuration}
As shown in Code Listing 1, there are a few configurations for CHOPT other than defining hyperparameter space.
It is necessary to define the goal of CHOPT, so that it can compare many models and find the best performing one.
Users can define the goal of a task by defining \textit{measure} and \textit{order}.
For example, we can select either top-1 accuracy or loss on the validation set as the \textit{measure}, and set descending or ascending as the \textit{order} respectively by their focus.  

As we mentioned in section~\ref{sec:resource_management}, our framework supports early stopping.
Setting the checking interval for stopping is important because it controls the performance of early stopping.
Here, users can configure this interval by setting \textit{step}. Note that if step is -1, CHOPT session 
disables early stopping.

Every iterative optimization process has a termination condition to avoid from running infinitely.
We support three options for $\textit{termination}$: $\textit{time}$, $\textit{max\_session\_number}$, and $\textit{performance\_threshold}$.
When multiple conditions are used, our framework stops as soon as it reaches the first condition. 

To choose a CHOPT algorithm we can configure it by \textit{tune}.
In this paper, we support three algorithms: random search with or without early stopping, population based training~\cite{jaderberg2017population}, and hyperband~\cite{li2017hyperband}.
Each algorithm requires different set of parameters.
For instance, population based training requires two functions to exploit to compare models and explore new search space while hyperband requires resource limit. 

%% file: 4_automlvis.tex
\subsection{Web-Based Analytic Environment}~\label{sec:automlvis}
Although hyperparameter optimization framework enables users to train hundreds,
thousands of models in parallel, if is difficult to analyze, results can be
useless.  CHOPT provides powerful visualization tools that allows users to
monitor, analyze the results to get constructive insights.  In this section, we
describe our visual tool's design goals, implementation and features.

\subsubsection{Design Goals}
\begin{itemize}
\item \textit{Scalable representation}: illustrate tuning results in a scalable way without loss of information even user adds hyperparameter more and more.
\item \textit{Compatibility environment}: integrate multiple results although they have different configurations such as the number of hyperparameter sets, tuning algorithm, and others.
\item \textit{Fine-tuning Method}: provide interface that simply re-configurate the hyperparameter spaces.
for searching unexplored or shrinking ranges.
\item \textit{Expand Hyperparameter Dimension}: allow user to add a new hyperparameter to be tuned.
\item \textit{Model Analyze}: serve many features for model analysis.
\end{itemize}

\subsubsection{Scalable Hyperparameter Configuration Visualization}

For scalable representation, we use parallel coordinates visualization
~\cite{inselberg1987parallel, heinrich2013state}.  Parallel coordinates
visualization has an advantage in representing high-dimensional or multivariate
data in a 2D plane by arranging each dimension in parallel.  Additionally, the
visualization can provide overall distributions of datasets for each dimension.
Other CHOPT visualization systems also have used parallel coordinates
visualization to represent hyperparameter tuning results
~\cite{golovin2017google, liaw2018tune, tsirigotis2018orion}.

\begin{figure}[!h]
\centering
\includegraphics[width=0.45\textwidth]{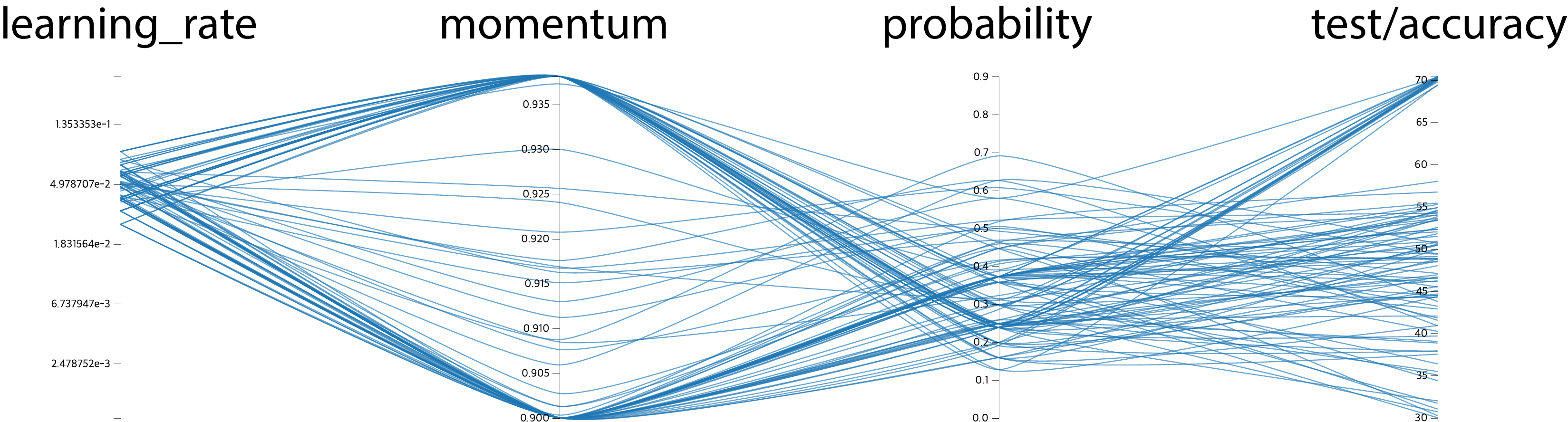}
\caption{An example of representation of hyperparameter tuning result on parallel coordinates visualization with on CIFAR100 random erasing model. Each axis represents each hyperparameter (except for last right axis for evaluation metric), and each line represents the trained model's hyperparameter configuration.}
\label{fig:parallel_coordinate_example}
\end{figure}

In the visualization as in Figure \ref{fig:parallel_coordinate_example}, each
line represents a machine learning model, NSML session, created by CHOPT
session, where each axis represents a hyperparameter such as learning rate,
activation, dropout rate, and more.  By arranging each hyperparameter in
parallel, the visualization can increase the number of hyperparameters and
present a number of hyperparameter configurations in a scalable way.  We expect
this visualization to help users analyze the relationships between
hyperparameters as well as analyzing results.

For the rest of design goals, we implement several features and they are
described in the following section.

\subsubsection{Visualization Feature List}
CHOPT visual tool serves many useful features to the user.
Users can monitor and analyze their models by interacting with visual tools.
We list up our selected features as follows.

\begin{itemize}
\item Model Selection and View
    \begin{itemize}
        \item Masking Top \textit{K} sessions: users can set thresholds and select models with the highest performances at once (Top at Figure~\ref{fig:desired model selection example}).
        \item Multiple range selection: users can select models with specific hyperparameter ranges
        (Bottom at Figure~\ref{fig:desired model selection example}).
        \item Highlighting axis: users can highlight lines with specific conditions to find relations between hyperparameters and model performance.
    \end{itemize}

\begin{figure}[h]
\centering
\includegraphics[width=0.45\textwidth]{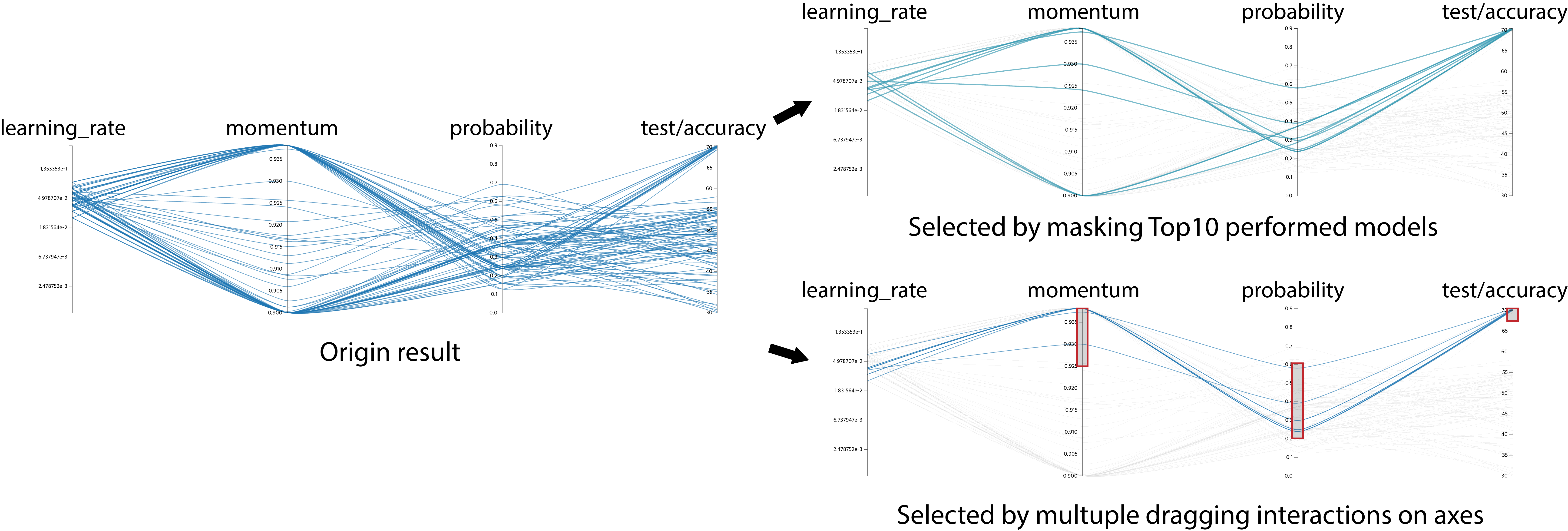}
\caption{An example of how select interested models on the parallel coordinates. Users can select the top performed \textit{K} models at once by one click and select the models within certain ranges by dragging the ranges on each axis.}
\label{fig:desired model selection example}
\end{figure}

\item Axis Control
    \begin{itemize}
        \item Filtering axis: if there are too many hyperparameters to analyze, users can specify the interested hyperparamters by setting the visibility of each axis.
        \item Controlling range scale: if there are too many lines to see the details on the parallel coordinates, users can easily adjust the scale of each axis by selecting the interested ranges on the density plot.
    \end{itemize}
\item Exhaustive Analysis
    \begin{itemize}
        \item Merging or switching interesting sessions: users can load and see the multiple autoML sessions, as well as merge and see the results simultaneously
        \item Scalar plot view: users can analyze the details of selected models such as loss values on each step or wall-time.
        \item Model summary view: users can easily see the precise values of the hyperparamters and the performance of selected models.
    \end{itemize}

\begin{figure}[h]
\centering
\includegraphics[width=0.45\textwidth]{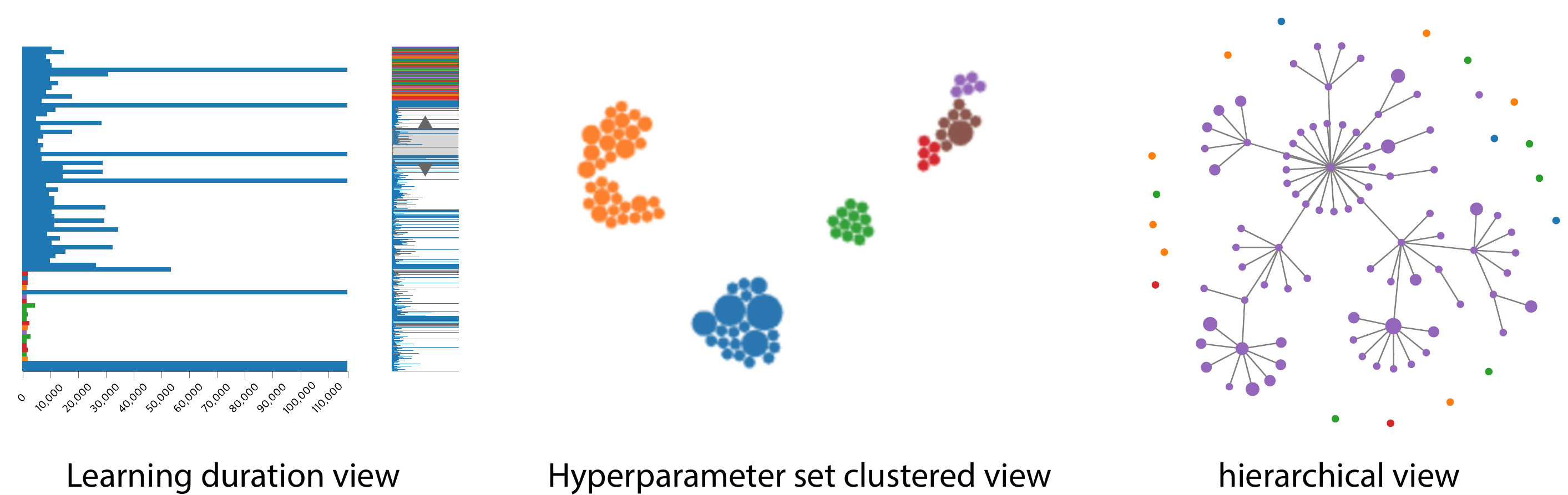}
\caption{An example of further analytic tools. Learning duration view is a bar plot for the models' learning duration; hyperparamter set clustered view is a t-SNE plot for the structural overview of created models; hierarchical view is a node-link diagram for the parent-child relation overview.}
\label{fig:analytics_insight}
\end{figure}

\item Further Analytic Tools
    \begin{itemize}
        \item Rerun Interface: users can change the configuration of a CHOPT session and submit a new one by selecting different hyperparameter ranges or adding hyperparameters on the parallel coordinates visualization.
        \item Parameter analytic view: users can select visualization type and parameters to see the distributions of the interested parameter or relation between two parameters via histogram or scatter plot. 
        \item Session overview: users can see the a summarized overview on hyperparameter configurations by selecting from learning duration, hierarchical, and clustered views.
    \end{itemize}
\end{itemize}

\begin{figure*}[!ht]
\centering
\includegraphics[width=0.75\textwidth]{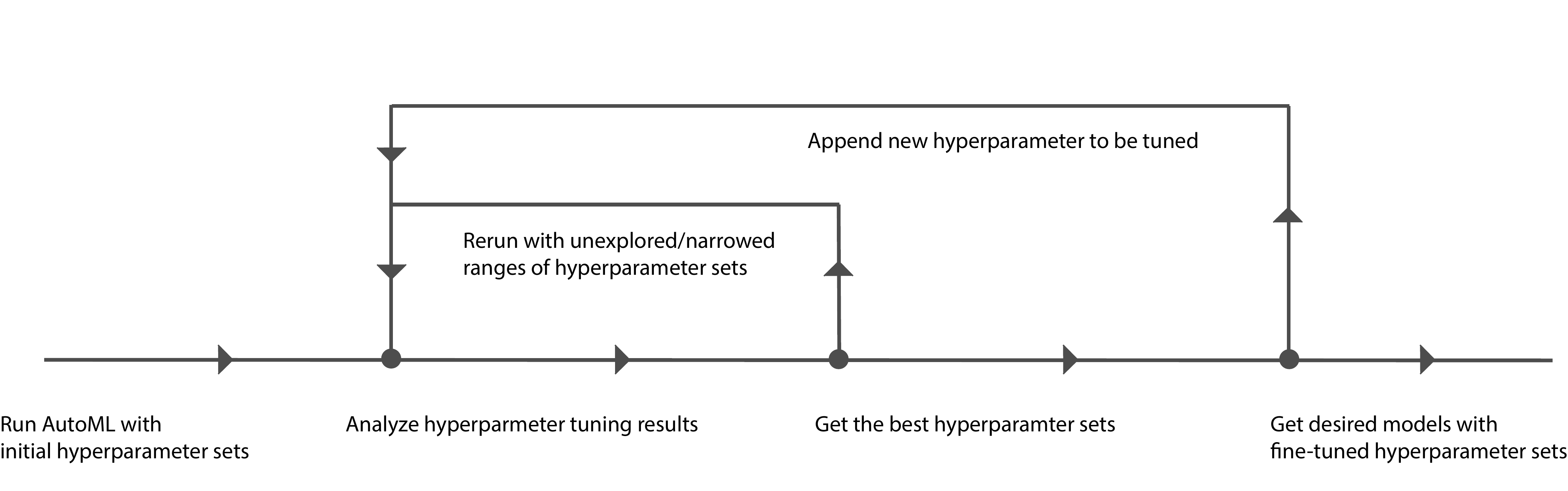}
\caption{Basic usage flow of hyperparameter tuning with CHOPT visualization}
\label{fig:basic_flow_of_vis}
\end{figure*}

\subsubsection{Fine Tuning with Analytic Visual Tool}~\label{sec:finetune}
When optimizing hyperparameters, the typical user would start with a small set of hyperparameters and add would incrementally expand the number of hyperparameters while optimizing.
This fine tuning process can be done with the following steps, which is also illustrated in Figure \ref{fig:basic_flow_of_vis}.

\begin{enumerate}
    \item Run CHOPT session with initial hyperparameter sets: 
    users run the CHOPT session with an initial configuration which contains hyperparameter sets to be tuned, ranges to be explored, and tuning method, and so on.
    
    \item Analyze hyperparameter tuning results: 
    users analyze the CHOPT results with a single session or multiple sessions according to whether the session is the initial one or not. We expect users will obtain insight into which values (for numerical) or types (for categorical) of hyperparameters affect the performance of the model at this Step.
    
    \item (Optional) Rerun with unexplored/narrowed ranges of hyperparamter sets: 
    users can select unexplored or narrowed ranges of the hyperparamter sets to make the CHOPT session explored the ranges based on the previous results. 
    In case of unexplored ranges, users may want to see the effect of other ranges to the performance of their models. 
    In case of narrowed ranges, they may want to find more precise values affecting the performance of the models. 
    Also, users can change the other configurations for the next CHOPT session such as tuning algorithm, step size for early stopping, population size, and so on. 
    Additionally, we expect that the users can combine tuning configuration based on what they aim at in this Step such as tuning algorithm, early stopping option, and etc.
        
    \item (Optional) Append new hyperparamter to be tuned:
    If users get an optimal combination of a given hyperparamter sets through a single CHOPT session or multiple sessions, they can append a new hyperparameter to be tuned which was a constant value in the previous sessions.
    
    \item Get desired models with fine-tuned hyperparamter sets
\end{enumerate}

%% file: 5_usecase.tex
\section{Practical use cases}

\begin{figure*}[ht!]
\centering
\includegraphics[width=0.97\textwidth]{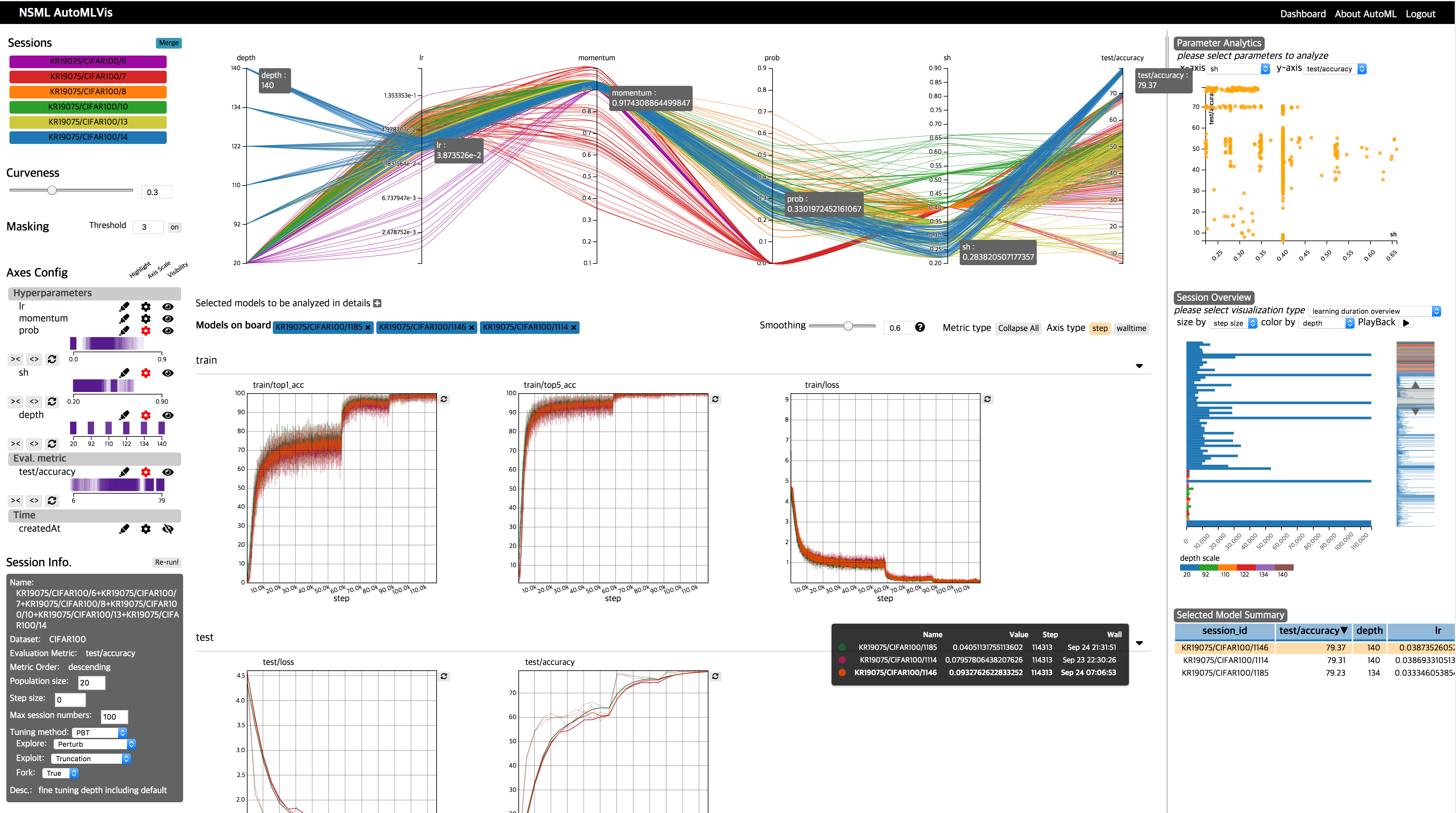}
\caption{An overview of CHOPT visualization system: the six different sessions with different hyperparameter configurations are represented on parallel coordinates, and the three top performed models are shown in details. The tooltips on parallel coordinates are activated by mouse hovered on the `train/loss' line plot on the middle of the figure.}
\label{fig:automlVis}
\end{figure*}

In this section, we describe how CHOPT visualization works with a
real-world example.
Figure \ref{fig:automlVis} shows an overview of CHOPT visualization with six CHOPT sessions with CIFAR100 dataset.

As described at the usage flow in Section ~\ref{sec:finetune}, we fine-tuned
six hyperparameters step-by-step. First, we run an initial CHOPT session
(purple color on the Figure \ref{fig:automlVis}) to tune `learning rate' with
other hyperparameters which not to be tuned.

For the next step, we selected the top ten models with our masking method
(figure \ref{fig:desired model selection example}) and obtained the optimal
ranges related to the selected models. With the optimized ranges of `learning
rate', we run a second CHOPT session with additional hyperparameter `momentum'
to be tuned (red color) so that CHOPT session explores the spaces with the optimal
range of `learning rate` and the initial range of `momentum'.

Subsequently, we obtained the optimized ranges that performed nicely and run
with additional hyperparameters, as same with the earlier steps.  With this
context, we gradually added other hyperparameters one by one and fine-tuned
each optimized range.

However, in the fifth step (yellow), we added `depth' hyperparameter and
obtained slightly improved performance with 70.54, and we found that the
experiment was quite biased by the early stopping because most models with
depths greater than twenty dropped very quickly.  Therefore, we rerun the
CHOPT sessions (blue color) without the early stopping method and obtained the
better results and the highest performed model with 79.37 percent accuracy,
which shown at the top-right side of the parallel coordinates in the Figure
\ref{fig:automlVis}. As we mentioned in Section, early stopping does not
guarantee better results than non early stopping results.

Table \ref{tab:fine_tuning_config} shows the changes of the configuration
(hyperparameter spaces) according to the progress of the fine tuning. Each
constant value means there was no exploration by the CHOPT.
Even if each CHOPT sessions has different configuration of hyperparamter
spaces to be tuned, our visualization allows users to integrate the sessions by
setting the constant value. For example, the `depth' hyperparamter was a
constant value of 20 in the all sessions (1-4th sessions in table) except for
the yellow and blue colored session (5-6th session in table) as shown in the
Figure \ref{fig:automlVis}.

\begin{table}[h]
\caption{Fine tuning results and configurations for each session in Figure \ref{fig:automlVis}}
\centering
\resizebox{\columnwidth}{!}{\begin{tabular}{l rrrrrrr}
\hline\hline
no. & Top Acc. & early stopped & learning rate & momentum & prob & sh & depth\\[1ex]
\hline\hline
1st & 69.62 & True & 0.001 - 0.2 & 0.9 & 0.0 & 0.4 & 20\\[1ex]
\hline
2nd & 69.78 & True & 0.0334 - 0.0868 & 0.1 - 0.999 & 0.0 & 0.4 & 20\\[1ex]
\hline
3rd & 70.4 & True & 0.0334 - 0.0868 & 0.9 - 0.9381 & 0.0 - 0.9 & 0.4 & 20\\[1ex]
\hline
4th & 70.36 & True & 0.0334 - 0.0868 & 0.9 - 0.9381 & 0.2378 - 0.579 & 0.2 - 0.9 & 20\\[1ex]
\hline
5th & 70.54 & True & 0.0266 - 0.0399 & 0.9 - 0.9381 & 0.2111 - 0.380 & 0.2237 - 0.3496 & [20, 92, 110, 122, 134, 140]\\[1ex]
\hline
6th & 79.37 & False & 0.0266 - 0.0399 & 0.9 - 0.9381 & 0.2111 - 0.380 & 0.2237 - 0.3496 & [20, 92, 110, 122, 134, 140]\\[1ex]
\hline\hline
\end{tabular}}
\label{tab:fine_tuning_config}
\end{table}

Meanwhile, the right side of the figure shows the further analysis of Automl
results and model summary table of the selected models.  The scatter plot
between `prob' hyperparamter and `test/accuracy' metric is visualized (right
top side of the figure), and it can be seen that the `sh' affects to high
performance values in the range of 0.22 to 0.35.

The horizontal bar plot (right middle) shows the learning duration of each
model. The x-axis represents the last learning step for each model.  In
practice, this plot can help users to find biased experiments because we found
that the fifth experiment had never explored at the depth with 140 by early
stopping.  The figure \ref{fig:early_stopping_barPlot} shows the difference
between fifth and sixth sessions which have same hyperparamter spaces but
different early stopping use. We found that the fifth session was biased to
depth with 20, and the depth 140 was not explored. On the other hand, the sixth
session was not biased.


The selected model summary table (the right-bottom side) is shown to provide
users to recognize the precise values of selected models at once such as
performance and hyperparamter configuration.

%% file: 6_evaluation.tex
\section{Evaluation}~\label{sec:evaluation}
In order to evaluate the performance of our system, we conduct experiments on two different tasks; Image classification and Question-Answering. For each experiment, 
we used CIFAR-100 dataset~\cite{cifar} and SQuAD 1.1~\cite{rajpurkar2016squad} dataset, respectively. 
Both evaluations show that our proposed framework finds better performance than 
performance reported in the papers.

\subsection{Image Classification and Question-Answering} 
Image classification and Question-Answering are well-known as the most challenging, and most popular tasks in deep learning community ~\cite{he2016deep, seo2016bidirectional}. There were huge breakthrough on convolutional neural networks (CNNs) for image classification after residual network ~\cite{he2016deep} and attention architectures for Question-Answering using SQuAD dataset, respectively. 
We test our CHOPT framework on various CNNs structures on the image classification task with CIFAR-100. We choose CIFAR-100 data because it is large enough to compare deeper model's performance comparing with mnist ~\cite{lecun1998mnist} or CIFAR-10, and simultaneously, it is small enough to train many models in a short time comparing with Imagenet ~\cite{deng2009imagenet}. We examine our framework with residual network (ResNet), wide residual network (WRN) ~\cite{zagoruyko2016wide}. In addition, we test on regularization method as well, specifically on data augmentation by Random Erasing (RE) ~\cite{zhong2017random} with ResNet and WRN to prove that our framework works on search space with high dimension. CHOPT is also evaluated on SQuAD 1.1 for Question-Answering. For experiments, we use BiDAF ~\cite{seo2016bidirectional}.
 In this experiment, we use random search with early stopping, population based training (PBT) and Hyperband while reporting the best result among these methods. 

\begin{table}[!htp]
\caption{Best top 1 accuracy (\%) reports obtained with CHOPT. Image classification (IC) and question-answering (QA) tasks were performed with various models. Bold indicates higher performance. ResNet~\cite{he2016deep}, WRN~\cite{zagoruyko2016wide}, RE~\cite{zhong2017random}}
\vskip 0.15in
\begin{center}
\begin{small}
\begin{sc}
\begin{tabular}{lcrr}
\toprule
Tasks & Models & References & CHOPT \\
\midrule
IC & ResNet  & 76.27 & \textbf{77.75} \\
\midrule
IC & WRN & 81.51 & \textbf{81.66} \\
\midrule
IC & ResNet with RE & 77.9 & \textbf{79.45} \\
\midrule
IC & WRN with RE & 82.27 & \textbf{83.1} \\
\midrule
QA & BiDAF & 77.3 & \textbf{77.93} \\
\bottomrule
\end{tabular}
\end{sc}
\end{small}
\end{center}
\vskip -0.1in
\label{tab:result_cifar}
\end{table}

Table~\ref{tab:result_cifar} summarizes the comparing results of the models reported by references and the best models searched by our CHOPT framework.
We use top-1 accuracy for measure, and in most cases, CHOPT succeeds in finding better-performing models than the references.

However, it may seem unfair to claim that CHOPT is better than the human-tuned model from the reference because of different model parameter sizes.
For instance, CHOPT's best model of WRN with RE in Table~\ref{tab:result_cifar} has 172.07M parameters while the best result from the reference contains only 36.54M. 
We tried one more experiment on WRN with RE while limiting the maximum number of parameters. 

\begin{table}[!htp]
\caption{Best model with parameter limit}
\vskip 0.15in
\begin{center}
\begin{small}
\begin{sc}
\begin{tabular}{lrr} 
\toprule
& Top-1 & \# of parameters \\
\midrule
Baseline  & 82.27\% & 36.54M \\
\midrule
CHOPT w/ constraint  & 82.41\% & 36.54M \\
\midrule
CHOPT w/o constraint  & 83.1\% & 172.07M \\
\bottomrule
\end{tabular}
\end{sc}
\end{small}
\end{center}
\vskip -0.1in
\label{tab:result_with_limit}
\end{table}

Table~\ref{tab:result_with_limit} shows the result of this experiment.
The best model by CHOPT is slightly better or at least same performance as human-tuned model even it is limited by the model size. 
Specifically, both models shared the same architecture of 28 depth and 10 widen factor, but they had different hyperparameters for regularization and optimization.

\subsection{Stop-and-Go}
Our early stopping method Stop-and-Go is one of the key features of our framework. In the best case scenario, early stopping can help users save both resources and time, while still finding the best model.
However, naive early stopping might fail and drop potentially good training sessions.
In this section, we will show how the early stopping affects performance and resource efficiency on CHOPT. 
We chose ResNet with RE as the target architecture for image classification on CIFAR-100 with the termination condition as 200 models for the experiment.
Each model is configured to run up to 300 epochs.
We used popular-based training (PBT) for all experiments with early stopping, and used random search without early stopping.

\begin{table}[h]
\caption{GPU time and performance by step size}
\vskip 0.15in
\begin{center}
\begin{small}
\begin{sc}
\begin{tabular}{lrr} 
\toprule
& GPU time & Top-1 \\[1ex]
\midrule
Without early stopping  & 60+ days & 79.75\% \\
\midrule
Large step size (25 epochs) & 22 days & 79.45\% \\
\midrule
Small step size (3 epochs) & 2 days & 77.42\% \\
\bottomrule
\end{tabular}
\end{sc}
\end{small}
\end{center}
\vskip -0.1in
\label{tab:gpu_efficiency}
\end{table}

Both resource efficiency and performance by step size are presented in Table~\ref{tab:gpu_efficiency}. 
We denote GPU time as the time taken if only one GPU is used for the entire CHOPT pipeline. 
Without early stopping, CHOPT can generate the best model among all algorithms including the human-tuned model (77.9\%), while taking more than 2 months of GPU time.
When we enable early stopping with small step size, it only takes 2 days of GPU time, but it cannot find a good enough model. 
However, if we put the right step size like the second row in the table, we can increase our GPU efficiency by more than a double, while obtaining similar performance without early stopping.

\begin{figure}[h]
\centering
\includegraphics[width=0.45\textwidth]{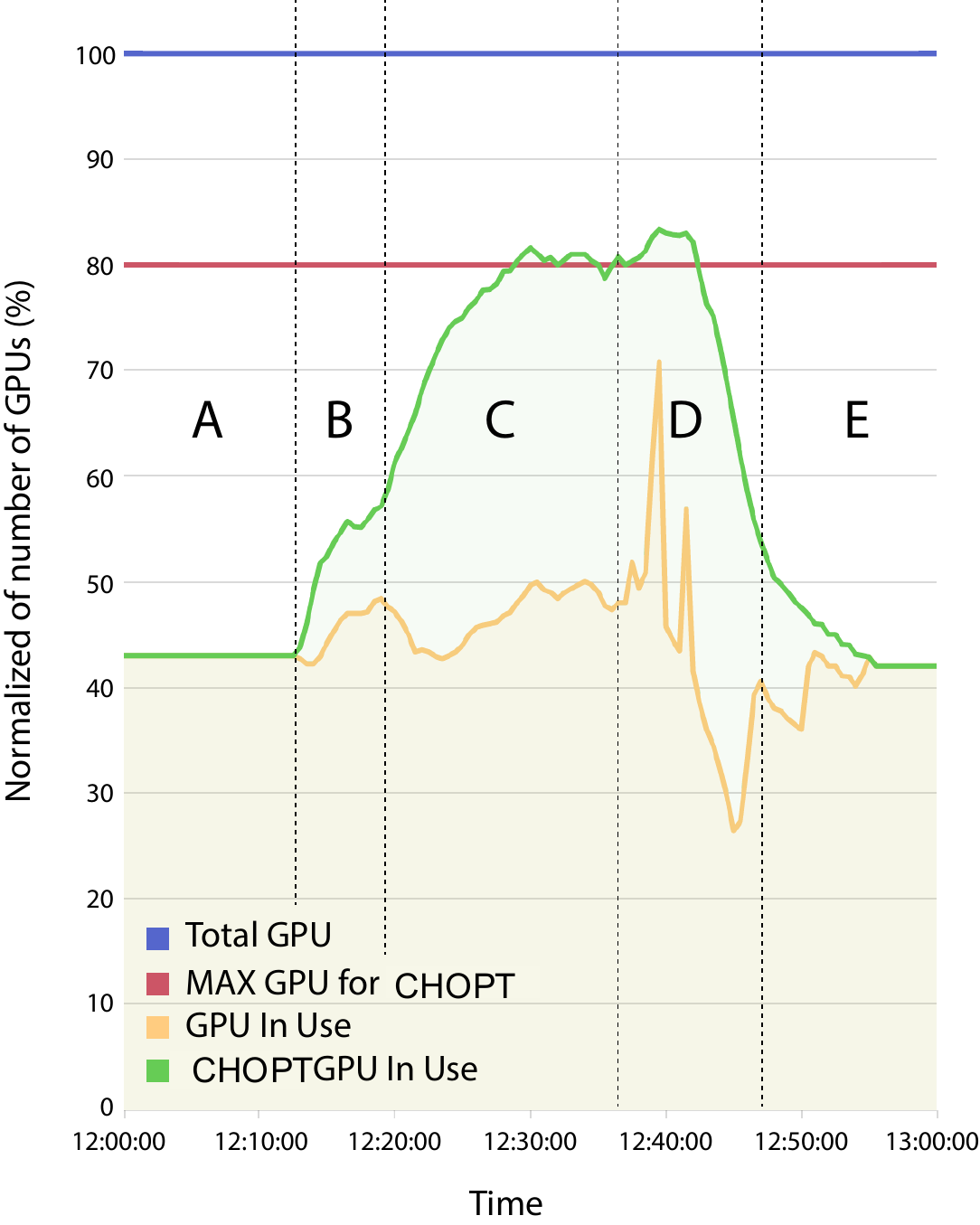}
\caption{Adaptive available GPUs control between NSML and CHOPT sessions}
\label{fig:gpu_utilize_graph}
\end{figure}

\begin{figure}[h]
\centering
\includegraphics[width=0.45\textwidth]{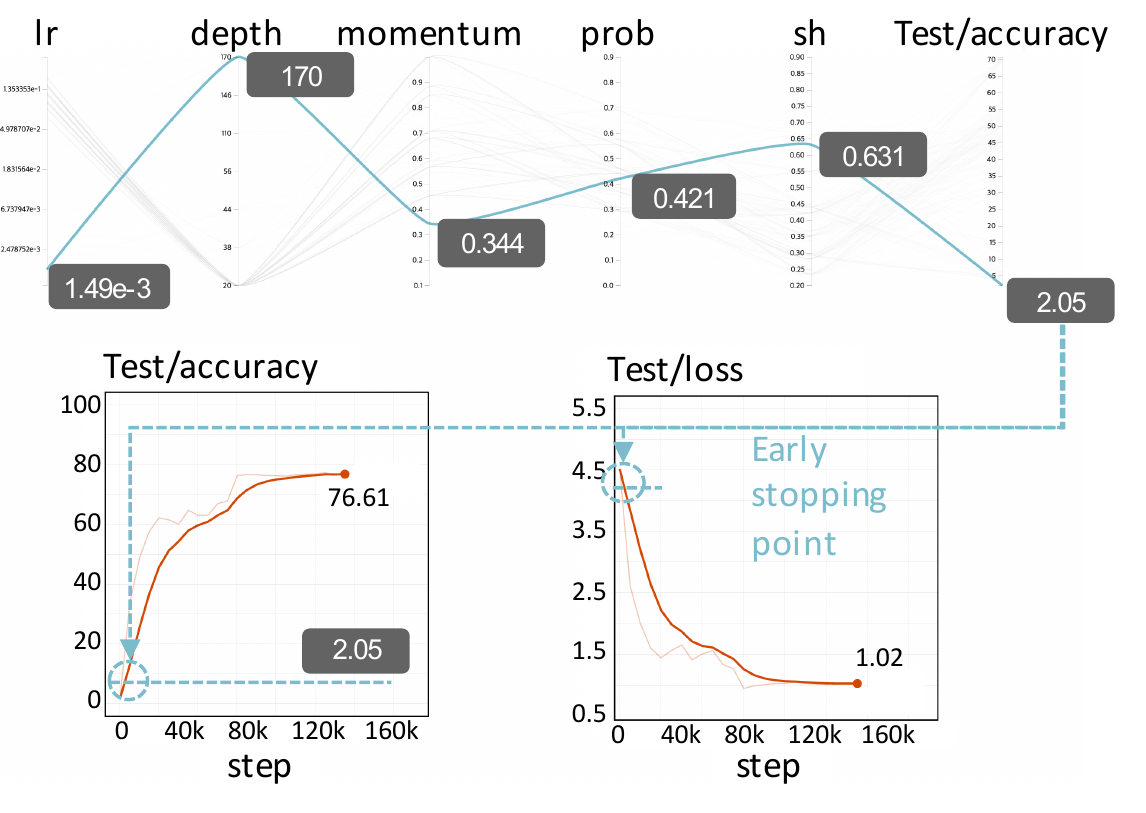}
\caption{Fully trained result of a revived early stopped model. Top: Hyperparameter configuration of early stopped model. Bottom: Fully trained result of early stopped model}
\label{fig:exp_cifar100_resume_bad}
\end{figure}

As another advantage, the stop-and-go feature allows us to get the same performance results in shorter time by assigning idle GPUs to CHOPT sessions. 
Figure~\ref{fig:gpu_utilize_graph} shows how proposed Stop-and-Go method
improves resource utilization. We divide time period into multiple zones,
A,B,C,D, and E for better explanation. In the time zone A, no CHOPT session is
running. So, green and yellow lines overlap. Then, several CHOPT sessions
start hyperparameter optimization in the time zone B. As a result, green line
goes up little bit. During the time zone C, master agent assigns idle GPUs to
CHOPT sessions to maximize the utilization of cluster and make tune much
faster since cluster is under-utilized. However, other users suddenly try to
use GPUs for their models in time zone D (Yellow line goes up). It turns out,
master agent takes GPUs from CHOPT session. Although it exceeds
maximum number of GPU for CHOPT but not that much. Last part of time zone D
and time zone E, CHOPT sessions almost finish optimizing hyperparameters so
green line goes down and finally overlap with yellow line.

By reviving stopped sessions from early stopping, the Stop-and-Go method sometimes improves performance of CHOPT. Figure~\ref{fig:exp_cifar100_resume_bad} presents one example of successful Stop-and-Go case.
As shown in the top side of Figure~\ref{fig:exp_cifar100_resume_bad}, the model was stopped by CHOPT at the early stage because its poor performance. 
However, after it got revived by Stop-and-Go, the model ended up with 76.61\% of top-1 accuracy.
Although it was not the best model CHOPT has found (77.42\%), considering that the results are competitive, we can say that Stop-and-Go can potentially save valuable hyperparameter configurations.

%% file: 7_conclusion.tex
\section{conclusion}~\label{sec:conclusion}

Although several HyperOpt methods have been proposed, many researchers still
suffer from optimizing hyperparameters due to the difficulty in usage and inconvenient interfaces. 
In this paper, we propose a novel cloud-based HyperOpt (CHOPT) framework that efficiently tackles flexible computing resources with supporting various HyperOpt algorithms.
In addition, we propose \textit{Stop-and-Go} to address the early stopping problem while maximizing utilization of shared-cluster resources. 
We presented our analytic visual tool and showed our system's capabilites with a practical use case of hyperparameter fine tuning. 
Finally, we demonstrate the performance of CHOPT with two real world tasks, image classification and question-answering.
The results have shown that our framework can find competitive hyperparameter configurations compared to previously reported results.

Although Stop-and-Go is an effective method, it is still hard to pick a session to revive among all early-stopped models.
For future work, we would like to find more effective policies other than using random selection methods.
In addition, we would like to extend our AutoML method by adding more criteria for our framework.
Currently CHOPT only uses for selecting the best model, but we would also like to find models which also have less number of parameters, are resource efficient, and train fast.